\documentclass{sig-alternate}
\pdfoutput=1

\usepackage{graphicx}
\usepackage{amsfonts}
\usepackage{amsmath}
\usepackage[small,tight]{subfigure}
\usepackage{balance}
\usepackage{multirow}
\usepackage{booktabs}

\begin{document}

\conferenceinfo{SIGSPATIAL}{'15, November 03-06, 2015, Bellevue, WA, USA}
\CopyrightYear{2015} 
\crdata{ISBN 978-1-4503-3967-4/15/11}  


\title{Land Use Classification using Convolutional Neural Networks Applied to Ground-Level Images}
\vspace{-3ex}
\numberofauthors{1} 
\author{
	Yi Zhu and Shawn Newsam\\
	\affaddr{Electrical Engineering \& Computer Science}\\
	\affaddr{University of California at Merced}\\
	\email{yzhu25,snewsam@ucmerced.edu}
}

\maketitle
\begin{abstract}
Land use mapping is a fundamental yet challenging task in geographic science. In contrast to land cover mapping, it is generally not possible using overhead imagery. The recent, explosive growth of online geo-referenced photo collections suggests an alternate approach to geographic knowledge discovery. In this work, we present a general framework that uses ground-level images from Flickr for land use mapping.

Our approach benefits from several novel aspects. First, we address the nosiness of the online photo collections, such as imprecise geolocation and uneven spatial distribution, by performing location and indoor/outdoor filtering, and semi-supervised dataset augmentation.  Our indoor/outdoor classifier achieves state-of-the-art performance on several benchmark datasets and approaches human-level accuracy. Second, we utilize high-level semantic image features extracted using deep learning, specifically convolutional neural networks, which allow us to achieve upwards of $76\%$ accuracy on a challenging eight class land use mapping problem.

\end{abstract}


\vspace{-1ex}

\begin{figure}[htb]
	\centering
	\includegraphics[width=0.9\linewidth,height=160pt,trim=0 0 0 0,clip]{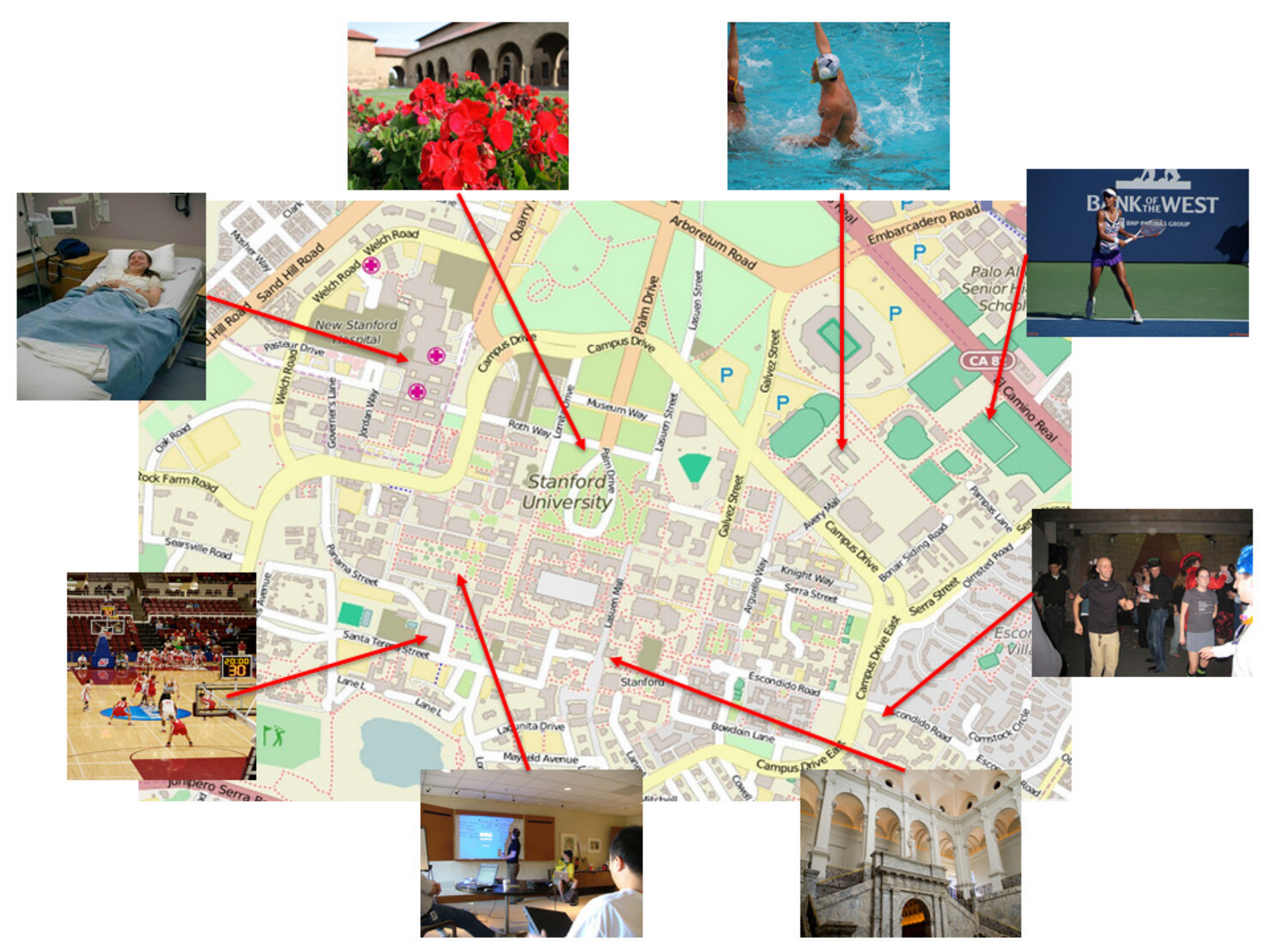}
	\vspace{-2ex}
	\caption{Sample ground-level images of Stanford University. What can these images tell us about the various land use classes found on the campus?\vspace{-3ex}}
	\label{fig:mapSample}
\end{figure}

\section{Introduction}
\label{sec:intro}

Innovative geographic knowledge discovery is becoming increasingly possible through analyzing large-scale online geotagged photo collections. 
For example, the authors in \cite{timeLapse15} recently introduced an approach called ``timelapse mining'' for synthesizing time-lapse videos of popular landmarks.  
Another interesting line of work that deals with image geolocation was initiated by Hays and Efros \cite{Haysim2gps08} and explores the problem of estimating the spatial coordinates of an image using a large dataset of images with known location. 
Other work, such as that of \cite{YanaiCulture09}, leverages large collections of geotagged images to discover spatially varying cultural differences among concepts such as ``wedding cake''. 
There are also individual projects such as the marvelous ``Geotaggers'' World Atlas by Erik Fischer, who aims to discover the world's most interesting places and the routes that people follow between them. 
There remain many opportunities for novel geographic knowledge discovery from this rich but complex data that is being acquired by millions of citizen sensors.


In this work, we focus on the challenging problem of mapping \emph{land use}. The salient contributions of our work include:
\vspace{-4ex}
\begin{itemize}
 \item We map a broader range of land use classes than previous work \cite{DanielLandUse12}. \vspace{-1.5ex}
 \item We utilize semantic image features learned by training convolutional neural networks, a form of deep learning, on a large collection of scene images.\vspace{-1.5ex}
 \item We develop an indoor/outdoor image classifier which achieves state-of-the-art performance. It helps correct for image location errors.\vspace{-1.5ex}
 \item Region shape files are used to further correct for image location errors as well as to create precise maps.\vspace{-1.5ex}
 \item A base set of training images is generated in an automated fashion and then augmented in a semi-supervised fashion to address class imbalance.\vspace{-1.5ex}
\end{itemize}


\section{Related Work}
\label{sec:relatedwork}
\noindent Our work has several lines of related research.

\noindent \textbf{Large-scale geotagged photo collections} Computer vision researchers have been leveraging large collections of geotagged photos for geographic discovery for around a decade. This includes mapping world phenomenon \cite{CrandallMapWorld09}, multimedia geo-localization \cite{Haysim2gps08}, landmark recognition \cite{3dModelSnavely08}, smart city \cite{smartUrban15}, land cover and land use classification \cite{DanielLandUse12,landcoverFlickr14} and ecological discovery. 
However, online photo collections are a noisy dataset, which presents a number of challenges in using them for geographic discovery. We address two of these challenges here. We use a semi-supervised learning framework to create a balanced training dataset, and we use shape files and indoor/outdoor classification to reduce geolocation error.



\noindent\textbf{Convolutional neural networks} Deep learning is advancing a number of pattern recognition and machine learning areas. Deep convolutional neural networks (CNNs) have resulted in often surprising performance gains in a range of computer vision problems \cite{AlexImageNet12}. These networks, whose training is made possible by the large-scale parallelization of graphical processing units (GPUs), consist of a number of convolutional layers which learn increasingly higher-level or semantic feature maps followed by one or more fully connected layers which perform classification. Indeed, in this paper, we use features learned from a large-scale scene classification task and achieve very promising results.



\noindent\textbf{Land cover and land use classification} Land cover classification is typically performed through the automated analysis of overhead imagery; e.g., the national land cover database (NLCD). However, the NLCD Level II (16 classes) overall accuracy for the 2006 map is only $78\%$ \cite{NCLD2006accuracy13}. Land use classification is even more difficult since it is often not possible from an overhead vantage point. However, the maps produced by land cover and land use classification are critical for a range of important societal problems.
Researchers have performed some initial investigation into using ground-level photo collections for land cover \cite{landcoverFlickr14} and land use \cite{DanielLandUse12} classification, but there remains significant opportunity to expand upon this initial work.

\begin{figure*}[htb]
	\centering
	\subfigure[]{\includegraphics[width=0.325\linewidth,height=120pt,trim=0 0 0 0,clip]{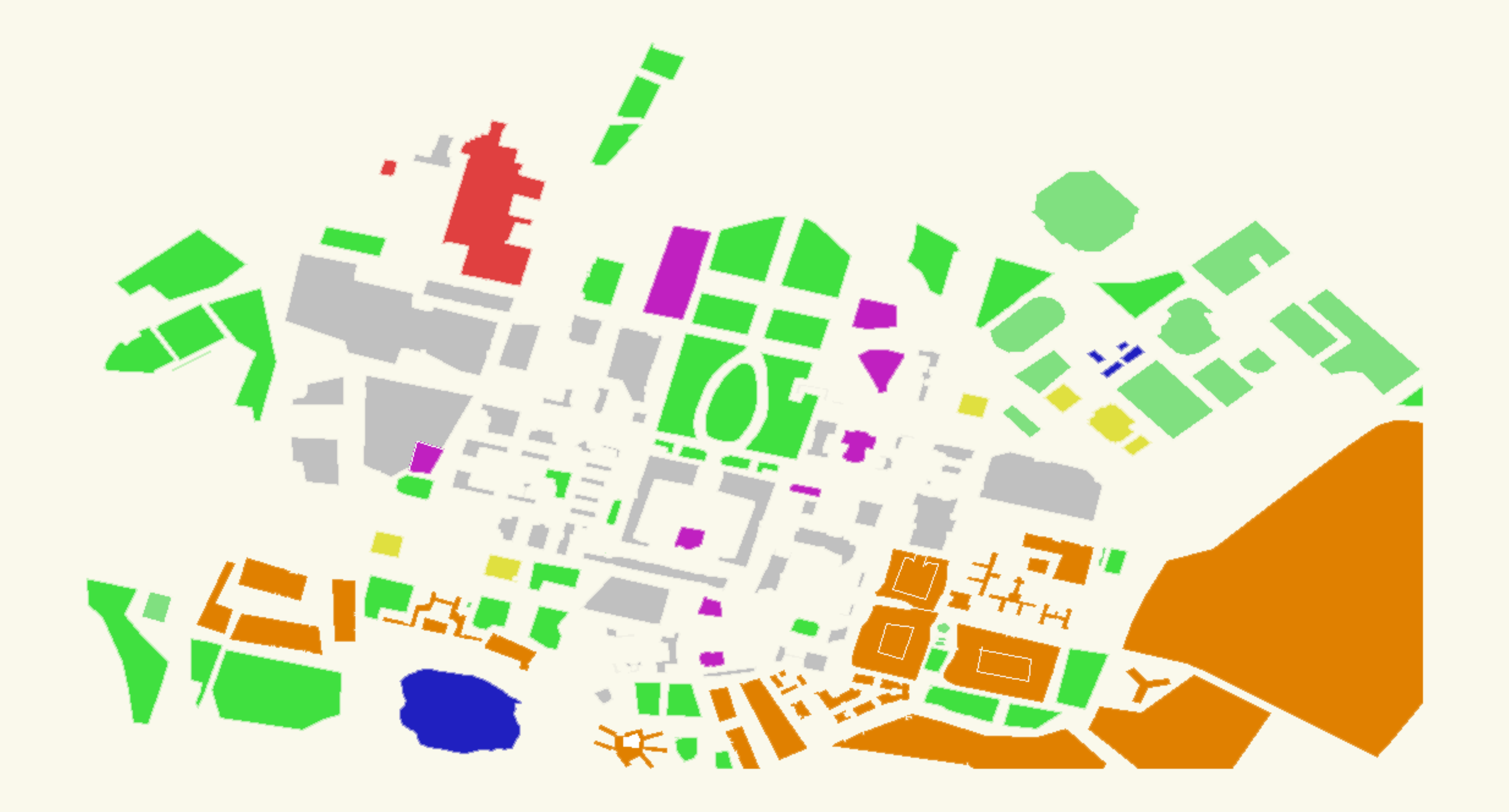}\label{fig:groundTruth8cat}}\hspace{0pt}
	\subfigure[]{\includegraphics[width=0.325\linewidth,height=120pt,trim=0 0 0 0,clip]{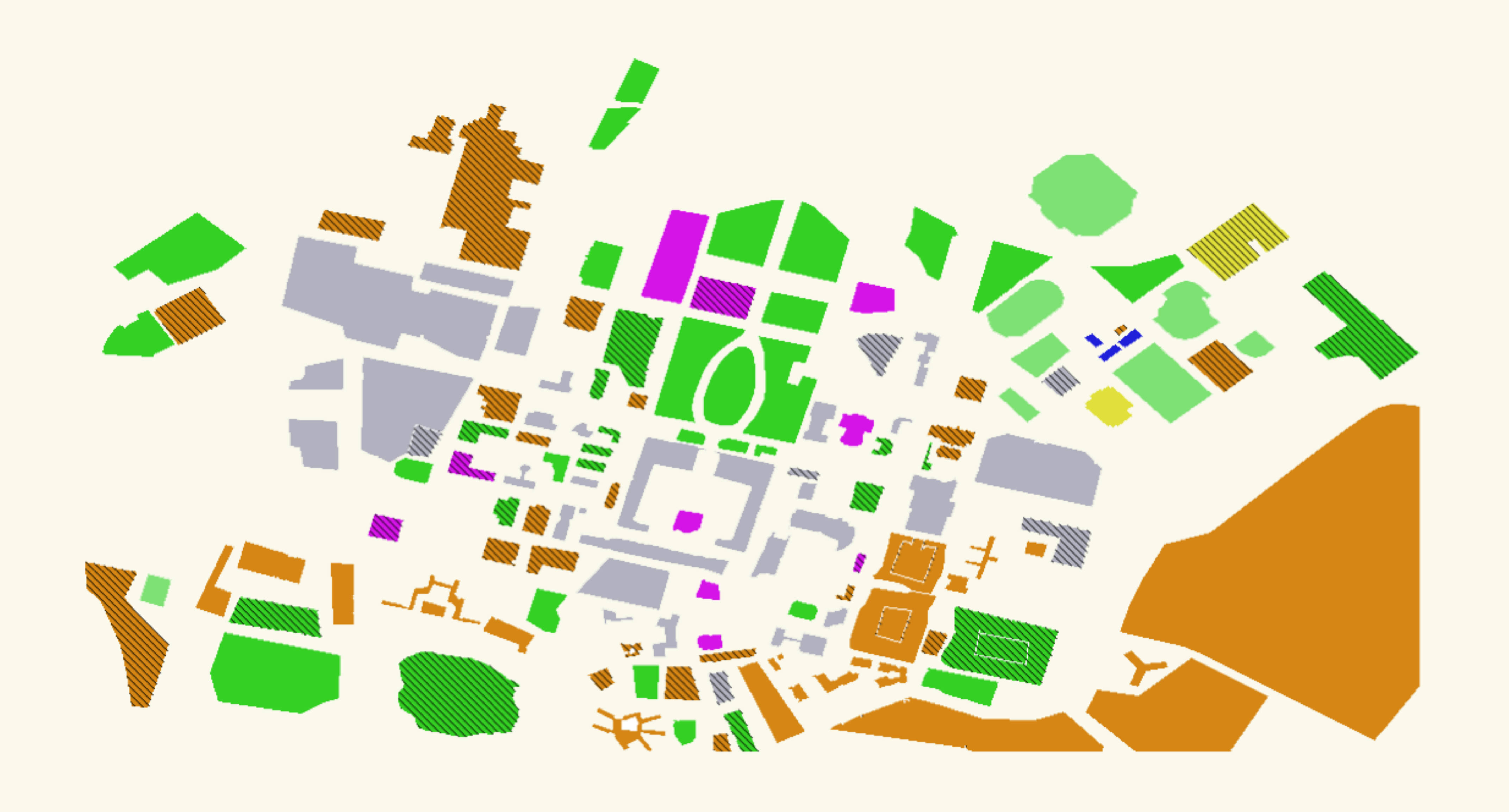}\label{fig:general1}}\hspace{0pt}
	\subfigure[]{\includegraphics[width=0.325\linewidth,height=120pt,trim=0 0 0 0,clip]{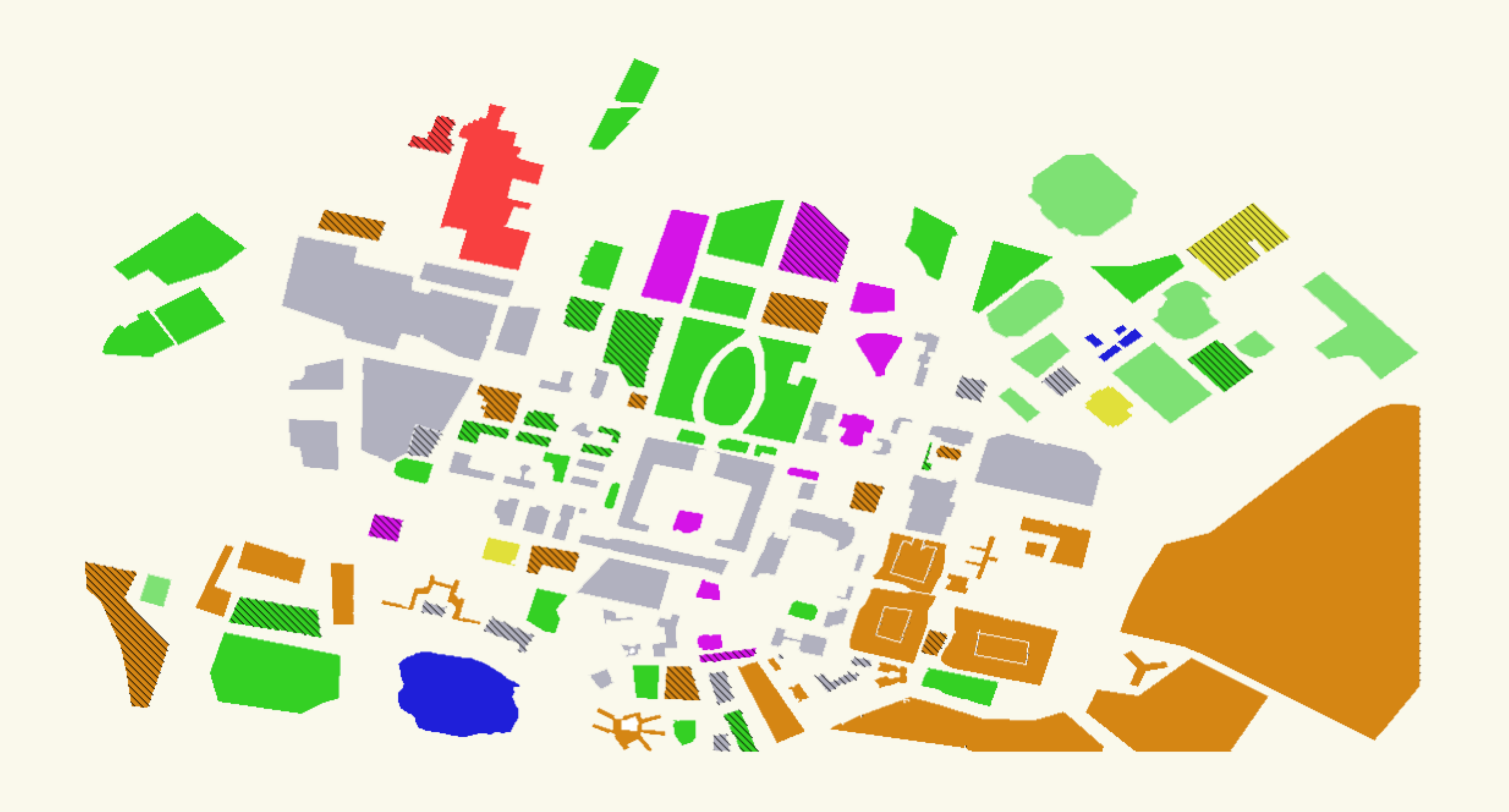}\label{fig:inout1}}\hspace{0pt}
	\vspace{-1ex}
	\caption{(a) Ground truth land use map. Gray, brown, red, dark green, yellow, light green, blue, and purple indicate study, residence, hospital, park, gym, playground, water, and theater, respectively. (b) Predicted land use maps without indoor/outdoor classification. (c) With indoor/outdoor classification.}
	\label{fig:experiment}
	\vspace{-2ex}
\end{figure*}

\begin{table*}
	\begin{center}
		\caption{Precision, recall, and F1 scores for image-level land use classification.\label{tab:recallPrecision8cat}}
		\vspace{0ex}
		\begin{tabular}{ l c c c c c c c c c }
			\toprule
			& Study & Residence & Hospital & Park & Gym & Playground & Water & Theater & \textbf{Average} \\
			\midrule
			\multicolumn{10}{c}{Without indoor/outdoor classification} \\
			Precision	& 0.5289 &	0.6282&	0.3145&	0.8168&	0.9676&	0.9653&	0.6638&	0.7823 & \textbf{0.7084}\\
			Recall	& 0.7952&	0.3596&	0.4762&	0.7165&	0.8627&	0.9772&	0.6814&	0.8128 & \textbf{0.7102}\\
			F1 score	& 0.6352&	0.4574&	0.3788&	0.7633&	0.9121&	0.9712&	0.6725&	0.7973 &\textbf{ 0.6985}\\
			\midrule
			\multicolumn{10}{c}{With indoor/outdoor classification} \\
			Precision	& 0.5581 &	0.6315&	0.4465&	0.8352&	0.9726&	0.9666&	0.7155&	0.7972 & \textbf{0.7404} \\
			Recall	& 0.8092&	0.3771&	0.5868&	0.7342&	0.8767&	0.9810&	0.7155&	0.8277 & \textbf{0.7385}\\
			F1 score	& 0.6606&	0.4721&	0.5071&	0.7815&	0.9222&	0.9737&	0.7155&	0.8121 & \textbf{0.7306}\\
			\bottomrule						
		\end{tabular}
	\end{center}
	\vspace{-2ex}
\end{table*} 

\vspace{-2ex}
\section{Land use classification}
\label{sec:classification}
\begin{figure}[htb]
	\centering
	\includegraphics[width=1.0\linewidth,trim=0 0 0 0,clip]{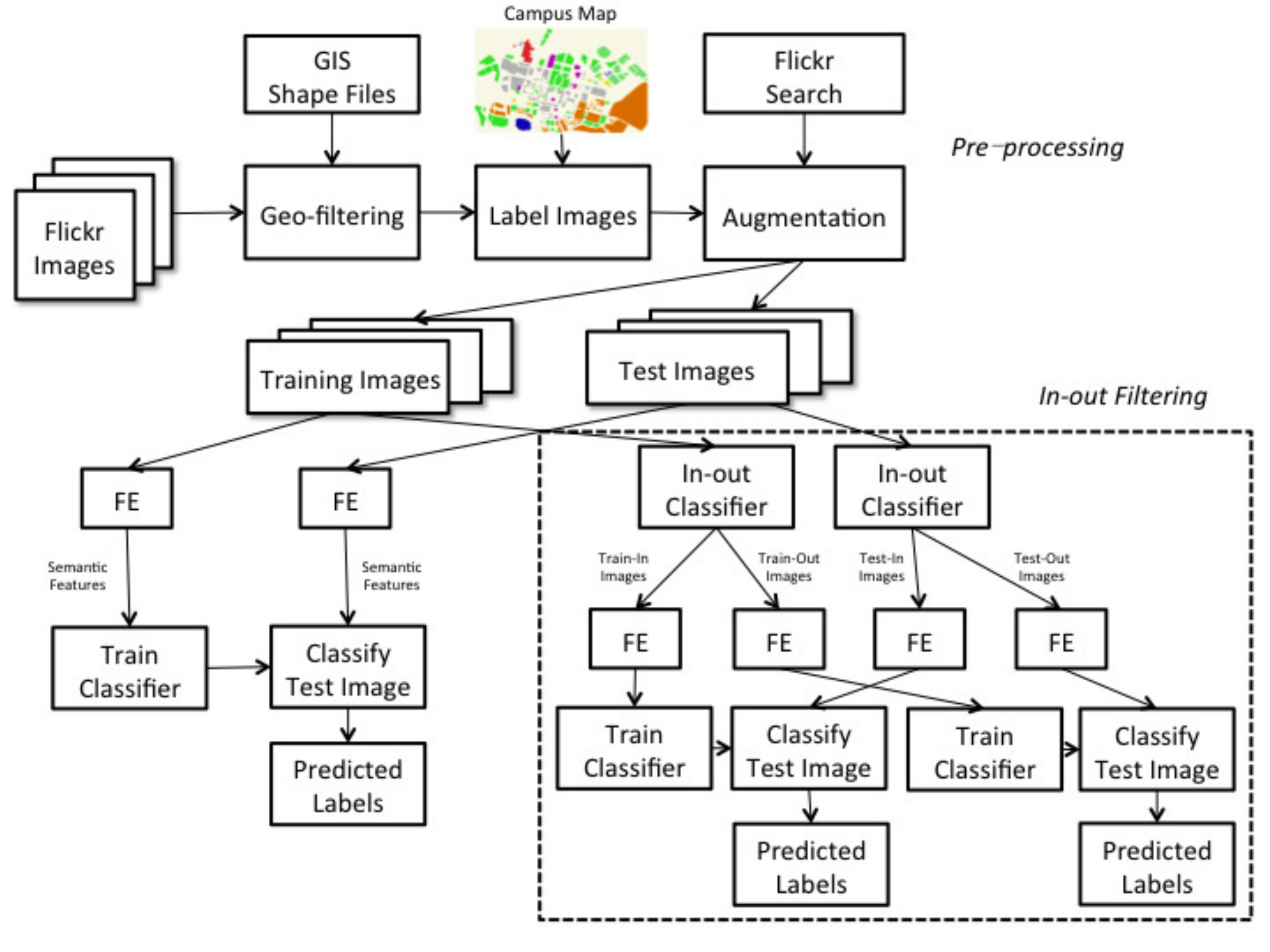}
	\vspace{-5ex}
	\caption{Workflow of the proposed framework. FE indicates image feature extraction. The dashed rectangle corresponds to the hierarchical model which incorporates indoor/outdoor classification. \vspace{0ex}}
	\label{fig:workflow}
	\vspace{-4ex}
\end{figure}

We focus on land use classification on a university campus since it represents a compact region containing a range of classes, and for which manually generating a ground truth is feasible. We consider eight land use classes on the Stanford University campus: \textbf{Study}, \textbf{Residence}, \textbf{Hospital}, \textbf{Park}, \textbf{Gym}, \textbf{Playground}, \textbf{Water} and \textbf{Theater}. Figure \ref{fig:mapSample} shows the map of the Stanford University campus from OpenStreetMap\footnote{https://www.openstreetmap.org/} and figure \ref{fig:groundTruth8cat} shows the ground truth land use map that we manually created. We use the Flickr API to download images located within the campus region, and each downloaded image is assigned a land use label according to its geographic location on the ground truth map. The workflow of our proposed framework is illustrated in figure \ref{fig:workflow} and will be discussed in the following sections.

\vspace{-1ex}
\subsection{Geo-Filtering with Shape Files}
\label{subsec:gis}
We use the polygonal outlines of the land use regions to filter noisy images and to produce more precise maps. Using the shape files as in figure \ref{fig:groundTruth8cat} has two benefits: 1). Filtering: We ignore the images which do not fall in one of the regions we want to classify. This removes a lot of noisy (unrelated) images and reduces our dataset from 79,658 images to 16,789; (2) Precision: The ground truth land use map in figure \ref{fig:groundTruth8cat} was generated using the shape files. It is very precise and could be published with very few modifications such as overlaying the street network. Compared with the tiling approach of previous work \cite{DanielLandUse12,NCLD2006accuracy13}, incorporating shape files results in maps that are significantly more geo-informative.
\vspace{-4ex}
\subsection{Dataset Augmentation}
\label{subsec:dataset}
We note, however, our campus dataset is quite unbalanced because of the uneven spatial distribution of Flickr photos. We thus propose a semi-supervised approach to augment the training set. This approach has three benefits: (1) it results in a balanced, richer training set; (2) it sets aside the vast majority of images with location for the test set; and (3) it is efficient and largely automated. The proposed procedure first randomly selects $20\%$ of each category in the geolocated dataset as the base training set, and the remaining geolocated images form the test set. Next, we perform a simple keyword search on Flickr for additional training images and augment the base training set with these auxiliary images which results in approximately 3,000 images per category. In addition to the actual category name, we perform searches using related keywords. We find the top retrievals are largely relevant and result in an effective augmented training set as demonstrated by our results below.

\vspace{-2ex}
\subsection{High-Level, Semantic Image Features}
\label{subsec:features}
Prior to recent advances in deep learning as applied to computer vision, image classification and scene understanding tasks largely utilized hand-crafted low-level features or combinations thereof. With the advent of large scale CNNs, we now have access to mid- to high-level image features that are derived in a data-driven fashion. These features have resulted in improved performance for a wide range of computer vision problems. Our work in this paper uses such features from a pretrained CNN model. The authors in \cite{zhouPlaces14} introduce a large scene-centric dataset called Places with over $7$ million labeled images. We use their released CNN-$205$ model\footnote{http://places.csail.mit.edu/} to extract our image features which are from layer seven as they perform better for our problem.

\vspace{4ex}
\subsection{Indoor/Outdoor Classification}
\label{subsec:inout}
While we cannot correct completely for images falling into the wrong region, we can use semantic-level signals, such whether an image is of an indoor or outdoor scene, to improve our land use classification.
%
%

The details of our hierarchical classification framework are as follows. \textit{Training phase}: 1) We first train an indoor/outdoor classifier leveraging the SUN dataset \cite{SUN2010}, then use this classifier to separate the training set into indoor and outdoor training sets (for each land use class); 2) We then train two eight-way land use classifiers, one for indoor images and another for outdoor images. \textit{Test phase}: 1) An incoming photo is first classified as indoor or outdoor; 2) We then use the corresponding classifier to assign the final land use label. To our knowledge, our indoor/outdoor classifier is shown to outperform state-of-the-art methods on several evaluation datasets. We achieve $98.64\%$ accuracy on the IITM-SCID2 dataset\footnote{http://www.cse.iitm.ac.in/$\sim$vplab/SCID/} and $99.15\%$ on the Fifteen Scene Categories dataset\footnote{http://www-cvr.ai.uiuc.edu/ponce$\_$grp/data/}. We also test with the latest SUN dataset ($908$ categories) to demonstrate the scalability and generalizability of our indoor/outdoor classifier. The best reported performance on this dataset is $91.15\%$ \cite{EDFinout14}, while we obtain $97.09\%$ on 66,406 test images.

\subsection{Experiments and Results}
\label{sec:experiment}
This section contains the experiments and results of our land use classification. We present image-level and region-level (map-level) performance. We first present results without indoor/outdoor classification and then with.

\subsubsection{Land Use Classification without In/Out}
\label{subsubsec:8cat} 
Here we train a single eight-way linear SVM classifier. The training set has 24,349 images (approximately 3,000 per class) and the test set has 13,435. The SVM is implemented using the LIBLINEAR package \cite{liblinear}, and the hyperparameter $C$ is determined using five-fold cross-validation. 

We apply the trained classifier to each test image. We label the map regions using majority vote of the contained images. A few regions do not contain any images. We do not attempt to label these regions. We obtain $77.55\%$ classification accuracy at the image-level. This is a solid result considering how noisy the Flickr images are with respect to location and content. This result is a testament to the high-level image features extracted using the CNNs as well as our training set augmentation. Table \ref{tab:recallPrecision8cat} lists the precision, recall, and F$1$ values for each category. Performance for the hospital class is the worst, possibly due in-part to the relatively low number of test images. Recall for the residence class is also quite low, possibly because it is quite comprehensive, including dorms, dining areas, party spaces, offices, indoor sports, etc.
We now label the map regions using the majority vote of the contained images. The result in figure~\ref{fig:general1} is shown to be very similar to the ground truth map in figure \ref{fig:groundTruth8cat}. $28$ of the $178$ regions contain no images and are ignored. We correctly classify $98$ of the remaining $150$ for an accuracy of $65.33\%$. 

\vspace{-1.5ex}
\subsubsection{Land Use Classification with In/Out}
\label{subsubsec:8catInOut}
We apply our hierarchical classification framework in which a test image is first classified as indoor or outdoor and then assigned a land use label. As described in section \ref{subsec:inout}, this helps compensate for geolocation error as well allows the final eight-way classification to be more discriminative since we now have two classifiers, one for indoor images and another for outdoor images. Indoor images will be less likely to be labeled as park, playground, or water. And, there should be fewer misclassifications of indoor images amongst the study, residence, hospital, gym, and theater classes.

The indoor land use classifier achieves $76.84\%$ image-level accuracy while the outdoor one achieves $80.85\%$. Taken together, this is an improvement over the non-hierarchical results. It also indicates that it is slightly easier to discriminate between the eight land use classes for the outdoor images than the indoor ones. The region-level results are displayed as a map in figure~\ref{fig:inout1}. Now only $36$ regions are misclassified for an accuracy of $76.00\%$. This is a big improvement over the non-hierarchical approach.

\vspace{-1ex}
\section{Ready-to-publish Maps}
\label{sec:publish}
\begin{figure}[htb]
	\centering
	\includegraphics[width=1.0\linewidth,trim=0 0 0 0,clip]{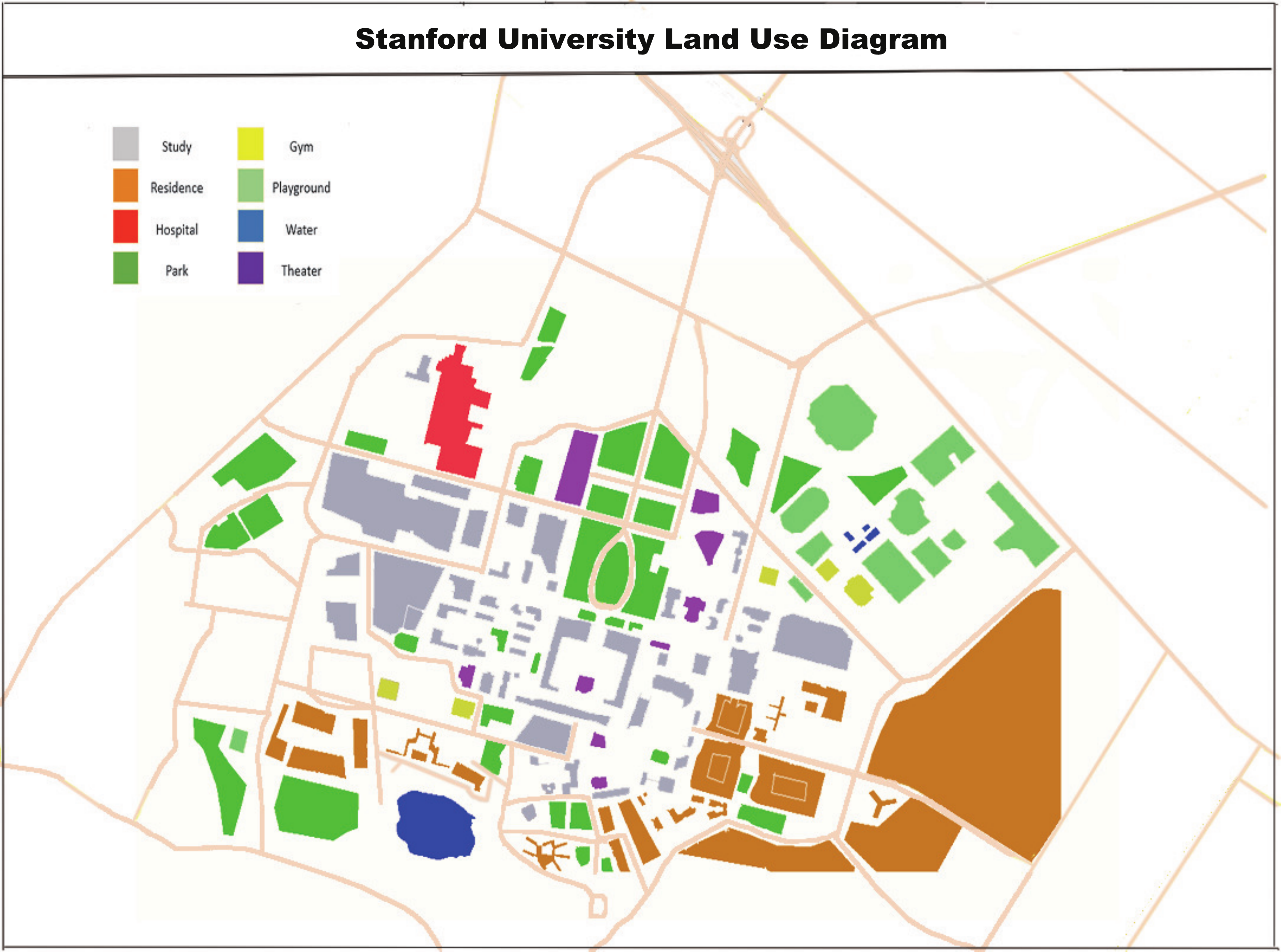}
	\vspace{-4ex}
	\caption{Our ready-to-publish land use map of the Stanford University region.\vspace{0ex}}
	\label{fig:publish}
	\vspace{-3ex}
\end{figure}

As mentioned earlier, the use of shape files in our approach results in spatially precise land use maps. With minor post-processing, our maps are ready to publish. To demonstrate, we overlay a road map of the Stanford University region from Wikimapia on our generated map. After adding a title and legend, and making a few minor revisions (e.g., changing the color of the background), the generated land use map is shown in figure \ref{fig:publish}.

\vspace{-1ex}
\section{Discussion}
\label{sec:discussion}
Our results above demonstrate that land use classification is possible using high-level image features extracted using CNNs from geolocated ground-level images. We believe our framework is ready to be applied to more complex and larger areas for which land use maps are not available or need to be updated, as long as there are a sufficient number of geolocated images.

A significant result of our work is showing that high-level, semantic image features extracted using pre-trained CNN models generalize well to related problems. This has very practical implications for researchers wanting to extract geographic information from georeferenced ground-level images. They do not need to buy expensive GPUs since these are only needed to train the models; feature extraction can be performed using standard computing hardware. They also do not need to go through the effort of configuring the models (which is often still through trial-and-error) nor spend the hours to days needed to train them. The fact that the high-level, semantic features allowed us to achieve such good performance using simple linear SVM classifiers further demonstrates the efficiency gain. The training time of our linear SVMs with over 20,000 samples of a 4,096-dimension feature vector is only $2.3$ seconds on a Dell workstation with a $3.3$ GHz quad-core CPU and 32GB RAM.

\vspace{-1ex}
\section{Conclusion}
\label{sec:conclusion}
We presented a framework for land use classification using geolocated ground-level images. Our approach maps a broader range of classes than previous work on this problem; uses high-level semantic image features extracted using CNNs, a form of deep learning; incorporates a novel, state-of-the-art indoor/outdoor classifier to help account for geolocation error; augments the training dataset in a semi-supervised fashion; and uses region shape files to produce precise maps. We achieve upwards of $76\%$ accuracy on a challenging eight-class land use classification problem.
\vspace{2ex}

\section{Acknowledgements}
This work was funded in part by a National Science Foundation CAREER grant, \#IIS-1150115.

\vspace{-1ex}
\bibliographystyle{abbrv}
\bibliography{landUse}  


\end{document}